\documentclass[11pt,a4paper]{article}
\usepackage[hyperref]{emnlp2020}
\usepackage{times}
\usepackage{latexsym}
\usepackage{booktabs}
\usepackage{graphicx}
\usepackage{xspace}
\usepackage{titlecaps}
\usepackage{multirow}

\usepackage{comment}
\usepackage{setspace}

\usepackage{microtype}
\usepackage{graphicx}
\usepackage{multirow}
\usepackage{booktabs}
\usepackage{tikz}

\definecolor{sc_color}{RGB}{12,7,134}
\definecolor{uwa_color}{RGB}{155,23,158}
\definecolor{propagated_color}{RGB}{236,120,83}

\usepackage{microtype}

\aclfinalcopy 

\title{Don't Neglect the Obvious:\\On the Role of Unambiguous Words in Word Sense Disambiguation}

\author{Daniel Loureiro \\
  LIAAD - INESC TEC \\
  University of Porto, Portugal \\
  \texttt{dloureiro@fc.up.pt} \\\And
  Jose Camacho-Collados \\
  School of Computer Science and Informatics \\
  Cardiff University, United Kingdom \\
  \texttt{camachocolladosj@cardiff.ac.uk} \\}
\date{}

\date{}

\begin{document}
\maketitle
\begin{abstract}
State-of-the-art methods for Word Sense Disambiguation (WSD) combine two different features: the power of pre-trained language models and a propagation method to extend the coverage of such models. This propagation is needed as current sense-annotated corpora lack coverage of many instances in the underlying sense inventory (usually WordNet). At the same time, unambiguous words make for a large portion of all words in WordNet, while being poorly covered in existing sense-annotated corpora. In this paper, we propose a simple method to provide annotations for most unambiguous words in a large corpus. We introduce the UWA (Unambiguous Word Annotations) dataset and show how a state-of-the-art propagation-based model can use it to extend the coverage and quality of its word sense embeddings by a significant margin, improving on its original results on WSD.
\end{abstract}

\section{Introduction}

There has been a lot of progress in word sense disambiguation (WSD) recently. This progress has been driven by two factors: (1) the introduction of large pre-trained Transformer-based language models and (2) propagation algorithms that extends the coverage of existing training sets. The gains due to pre-trained Neural Language Models (NLMs) such as BERT \cite{devlin-etal-2019-bert} have been outstanding, helping reach levels close to human performance when training data is available. These models are generally based on a nearest neighbours strategy, where each sense is represented by a vector, exploiting the contextualized embeddings of these NLMs \cite{melamud-etal-2016-context2vec,peters-etal-2018-deep,loureiro-jorge-2019-language}. However, training data for WSD is hard to obtain, and the most widely used training set nowadays, based on WordNet, dates back from the 90s \cite[SemCor]{Milleretal:93}. This lack of curated data produces the so-called knowledge-acquisition bottleneck \cite{gale1992method,navigli:09}.

However, there is a key source of information that has been neglected so far in existing sense-annotated corpora and propagation methods, which is the presence of unambiguous words from the underlying knowledge resource.
Strikingly, WordNet, which is known to be a comprehensive resource, is mostly composed of unambiguous entries (30k lemmas are ambiguous, compared to 116k unambiguous).
While the lack of unambiguous annotations does not have a direct effect in WSD, the fact that these unambiguous words are part of the same semantic network means they can have an effect on ambiguous words via standard propagation algorithms.
These propagation algorithms start from a seed of senses occurring in the training data (and therefore their embeddings can be directly computed) and then propagate to the whole sense inventory via the semantic network \cite{vial2018improving,loureiro-jorge-2019-language}.
Consequently, computing sense embeddings for unambiguous words can increase the number of seeds and improve the whole process.
Covering these unambiguous words, however, is not an arduous task, as unlabelled corpora may suffice.
We explore this hypothesis by labeling a large amount of unambiguous words in corpora extracted from the web, using WordNet as our reference sense inventory.
While we can certainly find usages of a word not covered by WordNet, we found that our approach can obtain accurate occurrences with simple heuristics.

The contribution of this paper is twofold.
First, we devise a simple methodology to construct UWA (Unambiguous Word Annotations), a large and, most importantly, diverse sense-annotated corpus that focuses on WordNet unambiguous words.
Second, we show that by leveraging UWA, we can significantly improve a state-of-the-art WSD model.

\section{Related Work}

The knowledge-acquisition bottleneck has been frequently addressed by automatically constructing sense-annotated corpora.
Recent works propose methods that exploit knowledge from Wikipedia, such as NASARI vectors \cite{camacho2016nasari}, for providing sense annotations for concepts and entities \cite{scarlini-etal-2019-just,pasini2019train}. In the case of \newcite{scarlini-etal-2019-just}, and similarly to \newcite{raganato-dellibovi-navigli:2016:IJCAI}, their method requires hyperlinks and category information from Wikipedia, hence not extensible to other kinds of corpora.\footnote{\newcite{pasini2020short} provide a more detailed overview of existing sense-annotated corpora.} 
Previous approaches relied on parallel corpora for two or more languages. The OMSTI corpus \cite{taghipour-ng-2015-one} was constructed by exploiting the alignments of an English-Chinese corpus. Similarly, \newcite{delli-bovi-etal-2017-eurosense} presented EuroSense, a multilingual sense-annotated corpus using the Europarl parallel corpus for 21 languages as reference. In contrast to these approaches, we focus on unambiguous senses and, therefore, are not constrained to only nouns, knowledge from Wikipedia, or a specific type of corpus.

Earlier works exploiting unambiguous words \cite{leacock-etal-1998-using,mihalcea-2002-bootstrapping,agirre-martinez-2004-unsupervised} and especially the subsequent extension by \citet{martinez_use_2008} are the most directly related to our paper. \citet{martinez_use_2008} retrieved example sentences with monosemous nouns from web search snippets and used them towards improved performance on WSD by leveraging WordNet relations. However, the WSD methods analyzed were sensitive to frequency bias, leading their collection effort to collect a large number of examples for fewer senses (and only nouns).
In contrast, our solution is designed for all monosemous words, retrieving examples from web texts instead of snippets, attaining performance gains with even a single example per word.

\section{Methodology}
\label{method}

In this section we first explain our method to construct a corpus with unambiguous word annotations (Section \ref{uwa}). Then, we explain current models based on language models for WSD (Section \ref{nlm}) and describe a propagation method to infer additional OOV sense representations (Section \ref{propagation}).

\subsection{Unambiguous Word Annotations (UWA)}
\label{uwa}

In order to properly test our hypothesis, we first require a sizable compilation of unambiguous words in context, particularly words that correspond to lemmas covered by WordNet. The extensiveness of WordNet means that most of its lemmas occur very rarely, and thus require processing large volumes of texts to achieve a high coverage. As such, in this work we develop the Unambiguous Word Annotations (UWA) corpus based on OpenWebText \cite{Gokaslan2019OpenWeb} and English Wikipedia (November 2019), processing over 53GB of texts from the web.

Each text is annotated for lemmas and part-of-speech using the  Stanford CoreNLP toolkit \cite{manning-etal-2014-stanford}. The annotations are filtered so that we only consider lemma/part-of-speech pairs that are present in WordNet, and correspond to a single sense (hence unambiguous), e.g., `keypad/noun'. Naturally, some lemma/part-of-speech pairs may have additional meanings not covered in WordNet. For example, in ``\textit{Inception} was a box-office hit.'', \textit{Inception} makes reference to a movie and not to the unambiguous word \textit{inception} from WordNet. To mitigate this issue, we applied Named Entity Recognition (NER) tagging, using spaCy \cite{spacy2}, to discard lemmas that are recognized as entities but do not correspond to an entity in their WordNet sense. To this end, we leverage the entity annotations of WordNet synsets available in BabelNet \cite{NavigliPonzetto:12aij}.
To keep the corpus at a reasonable size, we cutoff the maximum number of associated sentences (examples henceforth) per sense at 100.

\paragraph{Statistics.}
UWA covers a total of 98,494 senses, where 56.7\% have 100 examples, and 81.2\% have at least 10 examples. In Table \ref{tab:coverage} we show that UWA covers most senses for unambiguous words and, combined with SemCor, includes most senses in WordNet. This contrasts with other automatically-constructed datasets such as OMSTI \cite{taghipour-ng-2015-one} or T-o-M \cite{pasini2019train}. These sense-annotated corpora, not aimed specifically at unambiguous words, have limited coverage in this respect, as they are mainly composed of annotations for senses already available in SemCor.

\begin{table}[t]
\centering
\setlength{\tabcolsep}{3.0pt}
\resizebox{\columnwidth}{!}{
\begin{tabular}{r|cc|c||cc|c} \toprule
 & \multicolumn{2}{c|}{\textbf{\# Instances}} & \textbf{Avg} & \multicolumn{3}{c}{\textbf{Coverage (w/ SC)}}  \\
\textbf{Corpus} & \textbf{Amb}  & \textbf{Unamb} & \textbf{\# Exs} & \textbf{Amb}  & \textbf{Unamb} & \textbf{Total}  \\ \hline
SemCor & 198,153 & 27,883 & 6.8 & 26.2 & 7.4 & 16.1  \\
OMSTI & 909,830 & 1,304  & 244.7 & 26.8 & 7.4 & 16.4   \\
T-o-M & 719,888 & 114,580 & 152.4 & 28.5 & 7.5 & 17.2    \\ \hline
UWA(1) & 0 & 98,494 & 1.0 & 26.2 & 82.9 & 56.7    \\
UWA(10) & 0 & 804,861 & 8.8 & 26.2 & 82.9 & 56.7    \\
\multicolumn{1}{r|}{UWA(all)} & 0 & 6,111,453 & 54.1 & 26.2 & 82.9 & 56.7 \\ \bottomrule
\end{tabular}
}
\caption{Number of instances, average number of examples per word sense, and coverage percentage (including SemCor) of various sense-annotated corpora.}
\label{tab:coverage}
\end{table}

\subsection{Neural Language Models for WSD}
\label{nlm}

Recent NLMs, such as ELMo \cite{peters-etal-2018-deep} and BERT \cite{devlin-etal-2019-bert}, have been used with a high degree of success on WSD. They have been used differently depending on the nature of the disambiguation task: as feature providers for other neural architectures \cite{vial2019sensecompresswsd}, simple classifiers after fine-tuning \cite{wang2019superglue}, or as generators of contextual embeddings to be matched through nearest neighbours \cite[1NN]{melamud-etal-2016-context2vec,peters-etal-2018-deep,loureiro-jorge-2019-language,reif2019visualizing}. Our experiments in this paper will focus on improving the latter type of approach. In particular, we will investigate the state-of-the-art LMMS model \cite{loureiro-jorge-2019-language}. This model learns sense embeddings based on BERT states. These embeddings are then propagated through WordNet's ontology to infer additional senses, effectively providing a full coverage. 
While \citet{loureiro-jorge-2019-language} proposed variants of LMMS that combined propagation with gloss embeddings, or static embeddings, this paper is only concerned with the propagation method.

In our case, we essentially follow LMMS's layer pooling method to generate contextual embeddings for each sense occurrence in context (from a training set), and derive sense embeddings from the average of all corresponding contextual embeddings.

\subsection{Network Propagation for Full-Coverage}
\label{propagation}

The propagation method used in LMMS exploits the WordNet ontology to obtain a full coverage of sense embeddings from an initial set of embeddings based on a manually sense-annotated corpus like SemCor.
This method explores different abstraction levels represented in WordNet: sets of synonyms (synsets), Is-A relations (hypernyms) and categorical groupings (lexnames\footnote{Lexnames are also known as supersenses in the literature \cite{flekova-gurevych-2016-supersense,pilehvar-etal-2017-towards}.}).

Initial sense embeddings are first used to compute synset embeddings as the average of all corresponding senses (analogously to how sense embeddings are computed from contextual embeddings). From that point, missing senses are represented by their corresponding synset embeddings. The remaining unrepresented senses are inferred from their hypernym and lexname embeddings, computed by averaging their neighbour synset embeddings.
Note that this propagation process does not follow transitive relations in WordNet, i.e., a single synset's hypernym is considered, while the subsequent hypernyms along the root paths are ignored.

Since lexname embeddings can always be computed, this process can reach a full-coverage of WordNet starting with just the initial set of embeddings produced using SemCor. However, the set of SemCor embeddings only covers 16.1\% of WordNet, so many of the inferred representations are redundant and therefore not entirely meaningful.

\section{Evaluation}

For our experiments we are interested in verifying the impact of using UWA to improve WSD performance.
In particular, we test the unambiguous annotations of UWA as a complement of existing sense-annotated training data.
To this end, as explained in Section \ref{method}, we make use of the state-of-the-art WSD model LMMS \cite{loureiro-jorge-2019-language}. In addition to the original version using BERT, we also provide results with RoBERTa \cite{liu2019roberta} for completeness. We use the 24-layer models for both BERT and RoBERTa.\footnote{Commonly referred to as \textit{large} models.}

\subsection{Word Sense Disambiguation (WSD)}

\begin{table}[t]
\resizebox{\columnwidth}{!}{%
\begin{tabular}{@{}lrccccc|c@{}}
\toprule
\multicolumn{1}{l}{}          & \textbf{Corpus} & \textbf{SE-2} & \textbf{SE-3} & \textbf{SE07} & \textbf{SE13} & \textbf{SE15} & \textbf{ALL} \\ \midrule[\heavyrulewidth]
\multirow{6}{*}{\rotatebox[origin=c]{90}{\small{LMMS-BERT}}} & SC-noProp.         & 70.2          & 71.1          & 64.7          & 65.5          & 70.2          & 69.0         \\
                               & SC-only                                    & 75.5          & 74.2          & \textbf{66.8}          & 72.9          & 75.3          & 74.0         \\
                               & OMSTI          & 73.7          & 68.8          & 63.5          & \textbf{73.2}          & 74.8          & 71.9         \\
                               & T-o-M           & 69.9          & 66.1          & 62.4          & 64.8          & 74.2          & 67.9         \\
                               & UWA (1)         & 77.0          & \textbf{74.2}          & 66.2          & 73.1          & 75.4          & \textbf{74.5}         \\
                               & UWA (10)        & \textbf{77.3}          & 74.1          & 66.2          & 72.7          & \textbf{75.7}          & \textbf{74.5}         \\ \midrule
\multirow{6}{*}{\rotatebox[origin=c]{90}{\small{LMMS-RoBERTa}}} & SC-noProp.         & 70.7          & 70.6          & 66.7          & 65.1          & 70.5          & 69.2         \\
                               & SC-only         & 76.0          & 73.6          & \textbf{69.2}          & 72.3          & \textbf{75.9}          & 74.1         \\
                               & OMSTI           & 73.4          & 70.1          & 66.6          & 71.5          & 74.6          & 71.9         \\
                               & T-o-M           & 70.3          & 65.9          & 64.8          & 65.8          & 74.0          & 68.4         \\
                               & UWA (1)         & \textbf{77.8}          & 73.6          & 68.8          & 72.0          & 75.3          & 74.5         \\
                               & UWA (10)         & 77.6          & \textbf{73.7}          & 68.8          & \textbf{72.7}          & 75.3          & \textbf{74.6}         \\ \midrule 
                               \midrule 
\multirow{4}{*}{\rotatebox[origin=c]{90}{\small{SOTA}}}     & SC\textsuperscript{$\ddagger$}\textsubscript{LMMS+}     & 76.3          & 75.6          & 68.1          & 75.1          & 77.0          & 75.4         \\
                               &  SC\textsuperscript{$\dagger$}\textsubscript{\citeauthor{vial2019sensecompresswsd}}    & 76.6          & 76.9          & 69.0          & 73.8          & 75.4          & 75.4        \\
                               & SC\textsuperscript{$\ddagger$}\textsuperscript{$\dagger$}\textsubscript{EWISE}           & 73.8          & 71.1          & ~~67.3\textsuperscript{*}          & 69.4          & 74.5          & 71.8         \\
                               & SC\textsuperscript{$\ddagger$}\textsuperscript{$\dagger$}\textsubscript{\scriptsize{GlossBERT}}           & 77.7          & 75.2          & ~~72.5\textsuperscript{*}          & 76.1          & 80.4          & 77.0         \\
\bottomrule
\end{tabular}
}
\caption{F1 performance on the unified WSD evaluation framework.
All corpora marked are concatenated with SemCor (SC). SOTAs reported for reference but not directly comparable due to use of definitions (\textsuperscript{$\ddagger$}) or not using a 1NN approach (\textsuperscript{$\dagger$}).
All reported SOTAs are based on BERT trained on SC.
Results in datasets that were used as development are marked with \textsuperscript{*}.}

\label{tab:resultswsd}
\end{table}

Table \ref{tab:resultswsd} shows the WSD results on the standard evaluation framework of \newcite{raganato-etal-2017-word} for LMMS trained on the concatenation of SemCor and automatically-constructed corpora.
In the table we include UWA with two different maximum number of examples per unambiguous word, i.e., 1 and 10.
For comparison, we also include the results of EWISE \cite{kumar-etal-2019-zero} and GlossBERT \cite{huang-etal-2019-glossbert}, which attempt to overcome the limited coverage of SemCor by exploiting textual definitions.
As can be observed, the concatenation of our UWA corpus and SemCor provides the best overall results, regardless of the number of examples cut-off.
Perhaps surprisingly, our corpus is the only one that provides improvements over the baseline (SemCor-only). These improvements are statistically significant on the full test set (i.e. ALL) for both BERT and RoBERTa with p $<$ 0.0005, based on a t-test with respect to the accuracy scores (equal to F1 in this setting). This can be explained by the fact that our corpus is the only one that significantly extends the coverage of SemCor, as explained in Section \ref{uwa}.

\subsection{Uninformed Sense Matching (USM)}

In standard WSD benchmarks, models are given the advantage of knowing the pre-defined set of possible senses before-hand. This is because gold PoS tags and lemmas are provided in these datasets. However, to better understand how robust a 1NN WSD model is, we can test it in an uninformed setting, i.e., where PoS tags and lemmas are not given and the model does not have access to the list of candidate senses. Instead, the model has to match senses from the whole sense inventory, unconstrained. Therefore, in this Uninformed Sense Matching (USM) setting we can use information retrieval ranking metrics with the model predictions (i.e. MRR or P$@K$) in addition to the standard F1.
In line with the WSD results, Table \ref{tab:usm_comparison} shows that UWA also substantially improves performance in the USM setting when comparing against currently available alternatives.

\begin{table}[ht]
\centering
\resizebox{\columnwidth}{!}{%
\begin{tabular}{llccccccc} \toprule
\multirow{2}{*}{\textbf{Corpus}} &  & \multicolumn{3}{c}{\textbf{BERT}} & \textbf{} & \multicolumn{3}{c}{\textbf{RoBERTa}} \\ \cmidrule{3-5} \cmidrule{7-9} 
 &  & \textbf{F1} & \textbf{P@5} & \textbf{MRR} & \textbf{} & \textbf{F1} & \textbf{P@5} & \textbf{MRR} \\ \toprule
OMSTI &  & 50.2 & 66.0 & 57.5 &  & 44.1 & 59.9 & 51.7 \\
T-o-M &  & 45.8 & 62.1 & 53.3 &  & 42.1 & 60.7 & 50.2 \\
UWA (10) &  & \textbf{54.9} & \textbf{74.1} & \textbf{63.5} &  & \textbf{62.1} & \textbf{80.2} & \textbf{70.1} \\ \bottomrule
\end{tabular}%
}
\caption{Performance comparison in the uninformed setting. Each corpus is concatenated with SemCor.}
\label{tab:usm_comparison}
\end{table}

\section{Analysis}

In this section, we provide an analysis based on the number of examples (Section \ref{numberex}) and a visualization of the embedding space (Section \ref{visualizationemb}).

\subsection{Number of Examples}
\label{numberex}

\begin{figure}[t]
\centering
\includegraphics[width=\columnwidth]{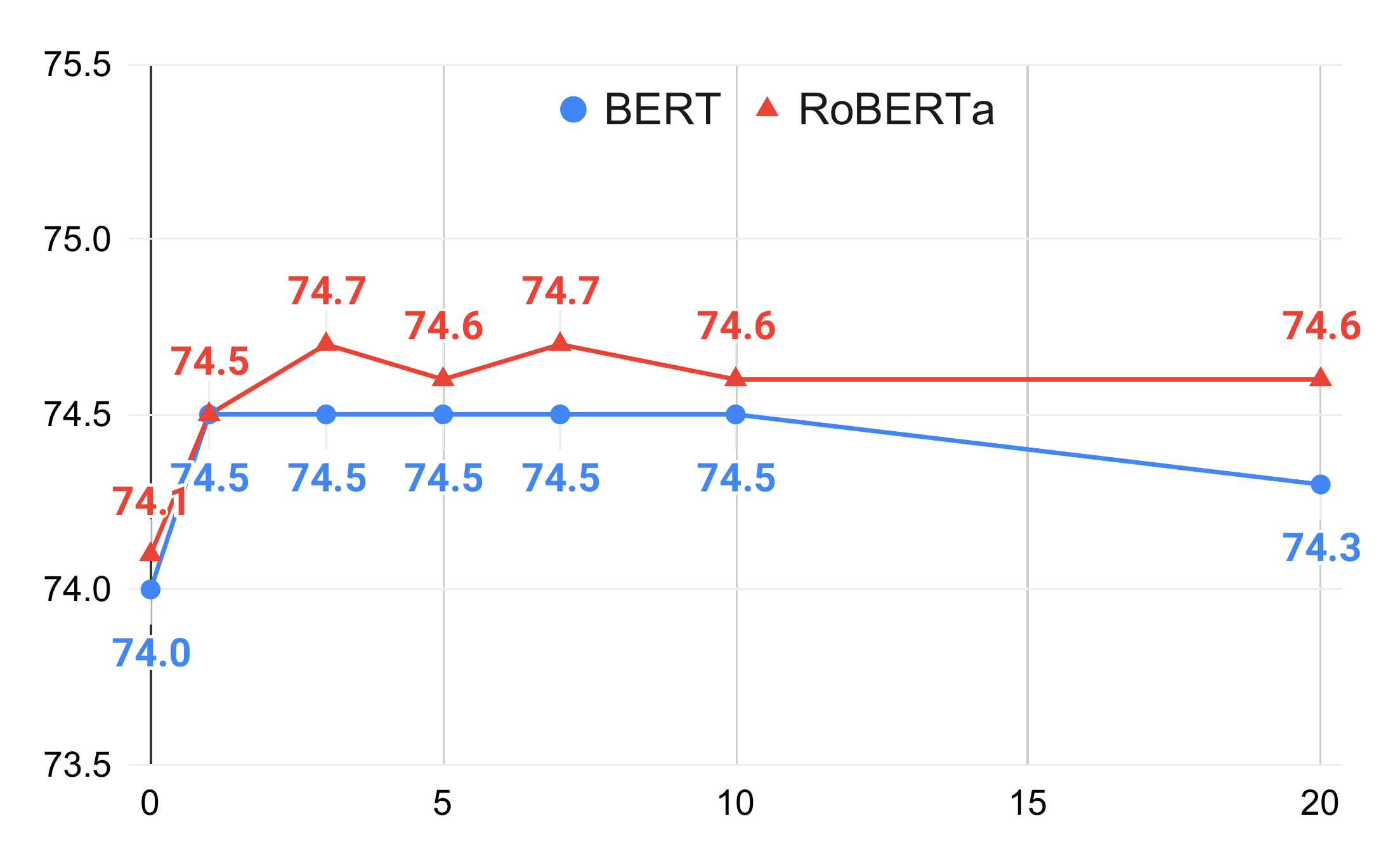}
\caption{WSD performance (F1 on the ALL test set) with different numbers of UWA examples.}
\label{fig:chart-examples}
\end{figure}

When compiling examples for learning sense representations, a natural question that arises is: how many examples are required to learn effective representations? The answer to this question can not only guide collection efforts, but also help clarify the requirements for learning effective representations in the simplest setting.
To that end, we analyse the impact of using different number of examples from UWA on LMMS's WSD and USM performance.
In Figure \ref{fig:chart-examples}, we show the WSD performance trend using different number of examples per sense.
As can be seen, performance improves substantially with only one example, and then stops improving after just two examples.

Similarly to our findings for WSD, Table \ref{tab:results-examples} shows that a low number of examples, such as 2, already achieves the best overall results in the USM setting for BERT. Likewise, RoBERTa does not benefit from more than 5 examples.
More generally, in USM the differences with respect to SemCor are more marked in comparison to the regular WSD setting.
This is expected as the propagation algorithm has a stronger effect in this setting where all sense embeddings are considered.

\begin{table}[t]
\centering
\resizebox{\columnwidth}{!}{%
\begin{tabular}{rcccccc}
\toprule
 & \multicolumn{3}{c}{\textbf{BERT}} & \multicolumn{3}{c}{\textbf{RoBERTa}} \\ \cmidrule(lr){2-4} \cmidrule(lr){5-7}
\multicolumn{1}{r}{\textbf{Corpus}} & \multicolumn{1}{c}{\textbf{F1}} & \multicolumn{1}{c}{\textbf{P@5}} & \multicolumn{1}{c}{\textbf{MRR}} & \multicolumn{1}{c}{\textbf{F1}} & \multicolumn{1}{c}{\textbf{P@5}} & \textbf{MRR} \\ \midrule[\heavyrulewidth]
\multicolumn{1}{r}{SemCor}           & \multicolumn{1}{c}{52.5}        & \multicolumn{1}{c}{67.1}         & \multicolumn{1}{c}{59.2}         & \multicolumn{1}{c}{58.0}        & \multicolumn{1}{c}{72.8}         & 64.7         \\ \midrule
\multicolumn{1}{r}{UWA (1)}  & \multicolumn{1}{c}{55.1}        & \multicolumn{1}{c}{74.1}         & \multicolumn{1}{c}{63.5}         & \multicolumn{1}{c}{61.3}        & \multicolumn{1}{c}{79.8}         & 69.5         \\
\multicolumn{1}{r}{UWA (2)}  & \multicolumn{1}{c}{\textbf{55.5}}        & \multicolumn{1}{c}{\textbf{74.6}}         & \multicolumn{1}{c}{\textbf{64.0}}         & \multicolumn{1}{c}{61.8}        & \multicolumn{1}{c}{\textbf{80.3}}         & 70.0         \\
\multicolumn{1}{r}{UWA (3)}  & \multicolumn{1}{c}{55.4}        & \multicolumn{1}{c}{74.5}         & \multicolumn{1}{c}{63.9}         & \multicolumn{1}{c}{61.9}        & \multicolumn{1}{c}{\textbf{80.3}}         & 70.0         \\
\multicolumn{1}{r}{UWA (5)}  & \multicolumn{1}{c}{55.4}        & \multicolumn{1}{c}{74.4}         & \multicolumn{1}{c}{63.8}         & \multicolumn{1}{c}{\textbf{62.1}}        & \multicolumn{1}{c}{\textbf{80.3}}         & \textbf{70.1}         \\
\multicolumn{1}{r}{UWA (7)}  & \multicolumn{1}{c}{55.2}        & \multicolumn{1}{c}{74.1}         & \multicolumn{1}{c}{63.7}         & \multicolumn{1}{c}{61.9}        & \multicolumn{1}{c}{\textbf{80.3}}         & 70.0         \\
\multicolumn{1}{r}{UWA (10)} & \multicolumn{1}{c}{54.9}        & \multicolumn{1}{c}{74.1}         & \multicolumn{1}{c}{63.5}         & \multicolumn{1}{c}{\textbf{62.1}}        & \multicolumn{1}{c}{80.2}         & \textbf{70.1}         \\
\multicolumn{1}{r}{UWA (20)} & \multicolumn{1}{c}{54.9}        & \multicolumn{1}{c}{73.7}         & \multicolumn{1}{c}{63.3}         & \multicolumn{1}{c}{\textbf{62.1}}        & \multicolumn{1}{c}{79.9}         & 70.0 \\
\bottomrule
\end{tabular}%
}
\caption{USM performance of the LMMS model using SemCor and UWA with different example thresholds. Models tested on the concatenation of all WSD datasets of \newcite{raganato-etal-2017-word}. As before, UWA is concatenated with SC in this experiment.}
\label{tab:results-examples}
\end{table}

\subsection{Visualization of the Embedding Space}
\label{visualizationemb}

\begin{figure}[t]
\centering
\includegraphics[width=0.75\columnwidth]{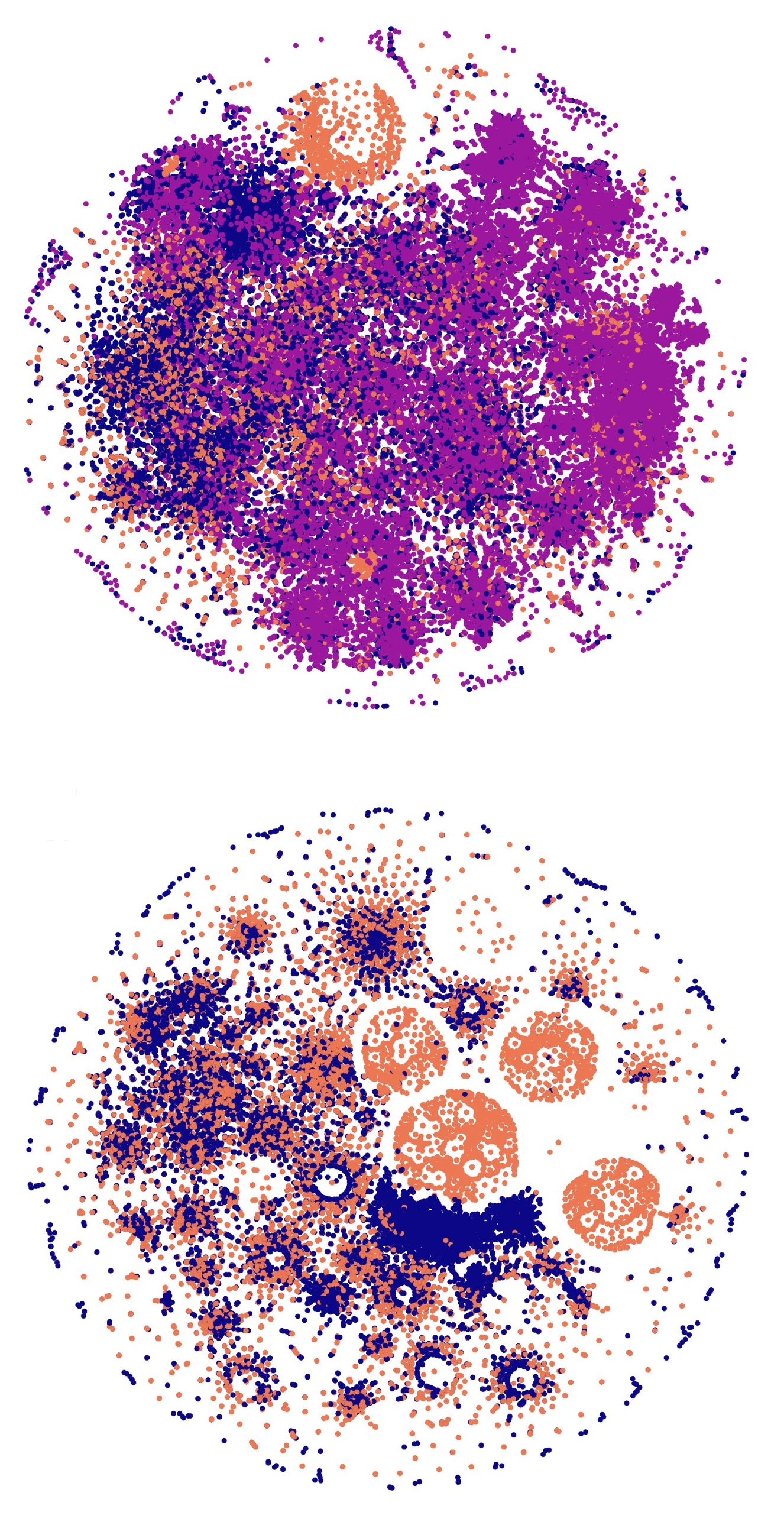}
\caption{T-SNE comparison of synset embeddings for whole WordNet learned from SC+UWA10 (top), or just SC (bottom). Colors represent source of annotations for embeddings (\tikz\draw[fill=sc_color] (0,0) circle (.7ex); SC \tikz\draw[fill=uwa_color] (0,0) circle (.7ex); UWA \tikz\draw[fill=propagated_color] (0,0) circle (.7ex); Propagation).}
\label{fig:tsne}
\end{figure}

The propagation method used in LMMS is designed to backoff to increasingly abstract representation levels, from synsets, to hypernyms, to supersenses (see Section 3.3 of the main paper). This naturally leads to a clustering effect, where many senses are represented with very similar, or equal, embeddings. In fact, we find that only 22\% of sense embeddings learned from SemCor, and propagated following LMMS, are actually unique (remaining are shared by two or more senses). The addition of UWA increases this percentage to 68\%.

To better understand this clustering effect, we used T-SNE  \cite{maaten2008visualizing} to visualize the WordNet synset embedding space.
In Figure \ref{fig:tsne} we show synset embeddings learned from the SemCor+UWA(10) dataset, and learned from SemCor alone, both based on RoBERTa.
While the same number of synset embeddings are learned in both cases, SemCor+UWA embeddings are better distributed across the vector space.
This, in turn, causes a substantial reduction of high-density clusters, which stand in opposition to a rich distributional representation of senses.\footnote{We share interactive visualizations focusing on each of the 45 supersense groups (e.g. \href{http://danlou.github.io/uwa/tsne/noun.communication_900px.html}{noun.communication}) from WordNet at our UWA release website.}

\section{Conclusion}

Unambiguous words are a surprisingly large portion of existing knowledge resources like WordNet. At the same time, their coverage in existing sense-annotated corpora is very limited. In this paper, we proposed a simple method which exploits sense annotations of unambiguous words from unlabeled corpora, thereby effectively extending existing sense-annotated corpora with low-effort. By leveraging a state-of-the-art BERT-based WSD system that propagates sense embeddings across WordNet, we have shown that these unambiguous words provide an excellent bridge to reach a wider range of OOV senses. This translates, in turn, into improving results for WSD. For future work it would be interesting to test these sense embeddings in a wider range of applications outside WSD.
Since the embedding space is clearly more diversified, as shown in Figure \ref{fig:tsne}, this may lead to improvements in other downstream tasks.

Moreover, one of the most surprising findings from this paper is that a single occurrence of OOV unambiguous words is enough to improve the performance of WSD models. This is relevant because (1) it is not always easy to retrieve a large number of examples for unambiguous words, and (2) it facilitates a cheaper manual verification, if required.

Finally, we openly release UWA, a large corpus annotated with unambiguous words, together improved BERT and RoBERTa-based sense embeddings, model predictions and visualizations at \url{http://danlou.github.io/uwa}.

\bibliography{anthology_subset,emnlp2020}
\bibliographystyle{acl_natbib}

\end{document}